\documentclass[11pt,a4paper]{article}
\usepackage[hyperref]{eacl2021}
\usepackage{times}
\usepackage{latexsym}

\usepackage{microtype}
\usepackage{booktabs}
\usepackage{graphicx}
\usepackage{multirow}
\usepackage{amsfonts}
\usepackage{tikz}
\usepackage{amsmath}
\usepackage{booktabs}
\usepackage{enumitem}
\usepackage{amssymb}
\usepackage{algpseudocode,algorithm,algorithmicx}
\usepackage{lipsum}

\newcommand\blankfootnote[1]{%
  \let\thefootnote\relax\footnotetext{#1}%
  \let\thefootnote\svthefootnote%
}

\aclfinalcopy 

\title{Progressively Pretrained Dense Corpus Index for\\ Open-Domain Question Answering}

\author{
 Wenhan Xiong$^\ast$,
 Hong Wang$^\ast$,
 William Yang Wang
\\ University of California, Santa Barbara\\
 \{xwhan, hongwang600, william\}@cs.ucsb.edu 
 }
\date{}

\begin{document}
\maketitle
\begin{abstract}

Commonly used information retrieval methods such as TF-IDF in open-domain question answering (QA) systems are insufficient to capture deep semantic matching that goes beyond lexical overlaps. Some recent studies consider the retrieval process as maximum inner product search (MIPS) using dense question and paragraph representations, achieving promising results on several information-seeking QA datasets. However, the pretraining of the dense vector representations is highly resource-demanding, \emph{e.g.}, requires a very large batch size and lots of training steps.
In this work, we propose a sample-efficient method to pretrain the paragraph encoder. First, instead of using heuristically created pseudo question-paragraph pairs for pretraining, we use an existing pretrained sequence-to-sequence model to build a strong question generator that creates high-quality pretraining data. Second, we propose a simple progressive pretraining algorithm to ensure the existence of effective negative samples in each batch. Across three open-domain QA datasets, our method consistently outperforms a strong dense retrieval baseline that uses 6 times more computation for training. On two of the datasets, our method achieves more than 4-point absolute improvement in terms of answer exact match.
\blankfootnote{$^\star$~Equal Contribution.}\blankfootnote{$^\_$~Our code is available at \url{https://github.com/xwhan/ProQA.git}.}
\end{abstract}

\section{Introduction}

With the promise of making the vast amount of information buried in text easily accessible via user-friendly natural language queries, the area of open-domain QA has attracted lots of attention in recent years. Existing open-domain QA systems are typically made of two essential components~\cite{chen-etal-2017-reading}. A \textit{retrieval module} first retrieves a compact set of paragraphs from the whole corpus (such as Wikipedia) that includes millions of documents. Then a \textit{reading module} is deployed to extract an answer span from the retrieved paragraphs.

Over the past few years, much of the progress in open-domain QA has been focusing on improving the reading module of the system, which only needs to process a small number of retrieved paragraphs. Specifically, improvements include stronger reading comprehension models~\cite{DBLP:conf/iclr/WangY0ZGCWKTC18,DBLP:journals/corr/abs-1902-01718,DBLP:journals/corr/abs-1912-09637,min-etal-2019-discrete} and paragraph reranking models~\cite{DBLP:journals/corr/abs-1709-00023,lin-etal-2018-denoising} that assign more accurate relevance scores to the retrieved paragraphs. However, the performance is still bounded by the retrieval modules, which simply rely on traditional IR methods such as TF-IDF or BM25~\cite{DBLP:journals/ftir/RobertsonZ09}. These methods retrieve text solely based on n-gram lexical overlap and can fail on cases when deep semantic matching is required and when there are no common lexicons between the question and the target paragraph. 
 
While neural models have proven effective at learning deep semantic matching between text pairs~\cite{bowman-etal-2015-large,DBLP:journals/corr/ParikhT0U16,DBLP:journals/corr/ChenZLWJ16,devlin-etal-2019-bert},
they usually require computing question-dependent paragraph encodings (\emph{i.e.}, the same paragraph will have different representations when considering different questions), which is formidable considering space constraints and retrieval efficiency in practice. More recent studies~\cite{lee-etal-2019-latent,chang2020pre,guu2020realm} show that such a dilemma can be resolved with large-scale matching-oriented pretraining. These approaches use separate encoders for questions and paragraphs and simply model the matching between the question and paragraph using inner products of the output vectors. Thus, these systems only need to encode all paragraphs in a question-agnostic fashion, and the resulted dense corpus index could be fixed and reused for all possible questions. While achieving significant improvements over the BM25 baseline across a set of information-seeking QA datasets, existing pretraining strategies are highly sample-inefficient and typically require a large batch size (up to thousands), such that diverse and effective negative question-paragraph pairs could be included in each batch. When using a small batch size in our experiments, the model ceases to improve after certain updates. Given that a 12G GPU can only store around 10 samples with the BERT-base architecture at training time, the wider usage of these methods to corpora with different domains (\emph{e.g.}, non-encyclopedic web documents or scientific publications) is hindered given modest GPU hardware.

In this work, we propose a simple and sample-efficient method for pretraining dense corpus representations. We achieve stronger open-domain QA performance compared to an existing method~\cite{lee-etal-2019-latent} that requires 6 times more computation at training time. Besides, our method uses a much smaller batch size and can be implemented with only a small number of GPUs, \emph{i.e.}, we use at most 4 TITAN RTX GPUs for all our experiments. In a nutshell, the proposed method first uses a pretrained sequence-to-sequence model to generate high-quality pretraining data instead of relying on heuristics to create pseudo question-paragraph pairs; for the training algorithm, we use clustering techniques to get effective negative samples for each pair and progressively update the clusters. Our method's efficacy is further validated through ablation studies, where we replicate existing methods that use the same amount of resources. For the downstream QA experiments, we carefully investigate different finetuning objectives and show the different configurations of the retrieval and span prediction losses have nontrivial effects on the final performance. We hope this analysis could save the efforts on trying out various finetuning strategies of future research that focus on improving the retrieval component of open-domain QA systems. 

The main contributions of this work include: 
\begin{itemize}[leftmargin=*]
     \item We show the possibility of pretraining an effective dense corpus index for open-domain QA with modest computation resources.
    
    \item Our data generation strategy demonstrates that pretrained language models are not only useful as plug-and-play contextual feature extractors: they could also be used as high-quality data generators for other pretraining tasks.
    
    \item We propose a clustering-based progressive training paradigm that improves the sample-efficiency of dense retrieval pretraining and can be easily incorporated into existing methods.
    
\end{itemize}

\section{Framework}

We begin by introducing the network architectures used in our retrieval and reading comprehension models. Next, we present how to generate high-quality question-paragraph pairs for pretraining and how we progressively train the retrieval model with effective negative instances. Finally, we show how to finetune the whole system for QA. 

\begin{figure*}[t]
\centering
\includegraphics[width=1.0\linewidth]{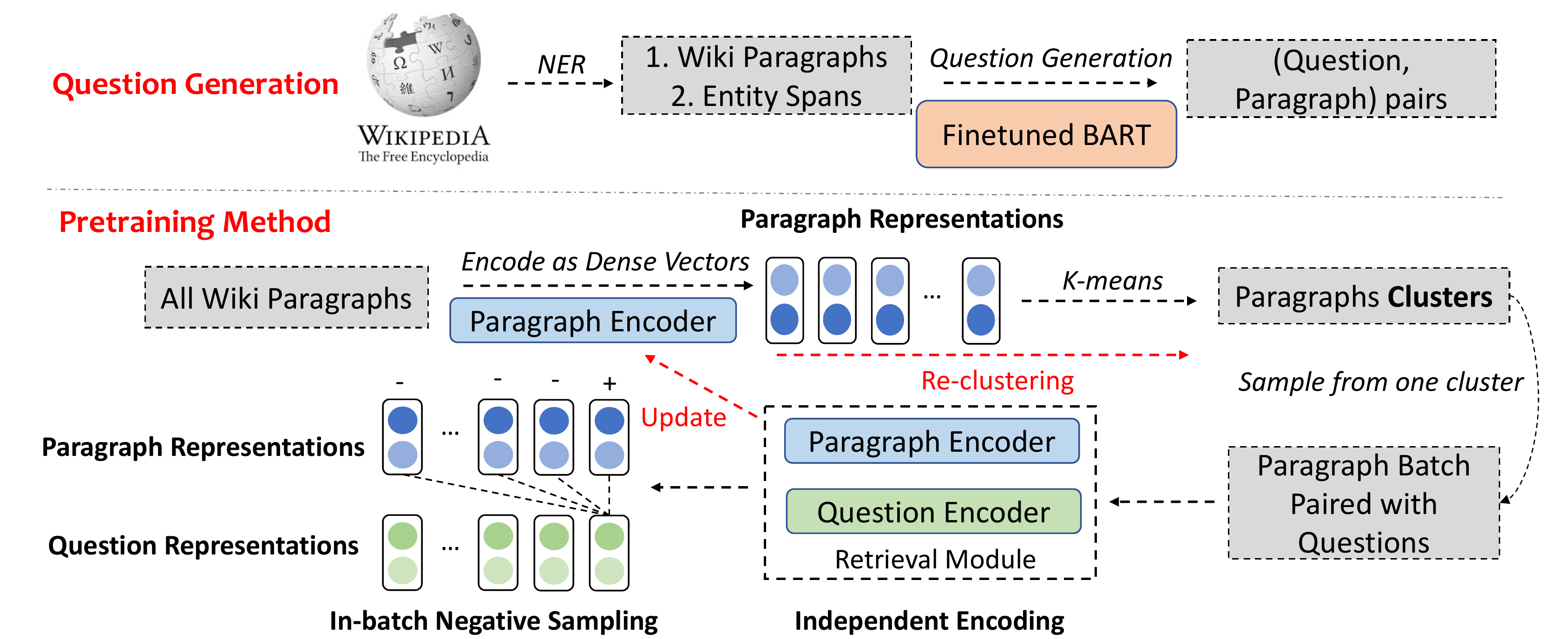}
\caption{An overview of the progressive pretraining approach.}
\label{fig:pretrain_method}
\end{figure*}

\subsection{Model Architectures}
\label{sec:model}

\paragraph{Notations} We introduce the following notations which will be used through our paper. The goal of open-domain QA is to find the answer derivation $(p,s)$ from a large text corpus $\mathcal{C}$ given a question $q$, where $p$ is an evidence paragraph and $s$ is a text span within $p$. The start and end token of $s$ are denoted as $\mathrm{START}(s)$ and $\mathrm{END}(s)$ respectively. We refer the retrieval module as $\mathrm{P_{\theta}}(p|q)$, with learnable parameters $\theta$. Similarly, we refer the reading comprehension module as $\mathrm{P}_{\phi}(s|p,q)$, which can be decomposed as $\mathrm{P}_{\phi}(\mathrm{START}(s)|p,q) \times \mathrm{P}_{\phi}(\mathrm{END}(s)|p,q)$. We use $D^k$ to represent the top-k paragraphs from the retrieval module; a subset of $D^{\ast} \in D^{k}$ represents the paragraphs in $D^k$ that cover the correct answer; for each paragraph  $p \in D^{\ast}$, we define $S^{\ast}_{p}$ as all the spans in $p$ that match the answer string. 

\paragraph{The Retrieval Module} 
We uses two isomorphic encoders to encode the questions and paragraphs, 
and the inner product of the output vectors is used as the matching score. The encoders are based on the BERT-base architecture. We add linear layers $\mathbf{W}_q\in\mathbb{R}^{768\times128}$ and $\mathbf{W}_p\in\mathbb{R}^{768\times128}$ above the final representations of the \textsc{[CLS]} token to derive the question and paragraph representations:
\begin{align*}
     &h_{q} =\mathbf{W}_q \mathrm{BERT}_{Q}(q)([\mathrm{CLS}]) \\
    &h_{p} =\mathbf{W}_p \mathrm{BERT}_{P}(p)([\mathrm{CLS}]), 
\end{align*} 
The matching score is modeled as $h_{q}^{\top} h_{p}$. Thus, the probability of selecting $p$ given $q$ is calculated as:
$$
\mathrm{P}_{\theta}(p|q)=\frac{e^{h_{q}^{\top} h_{p}}}{\sum_{p^{\prime} \in \mathcal{C}}e^{h_{q}^{\top} h_{p^{\prime}}}}.
$$
In practice, we only consider the top-k retrieved paragraphs $\mathcal{C}$ for normalization. 

\paragraph{The Reading Module} The architecture of our reading comprehension model is identical to the one in the original BERT paper~\cite{devlin-etal-2019-bert}. We use two independent linear layers to predict the start and end position of the answer span. At training time, when calculating the span probabilities, we apply the shared-normalization technique proposed by \citet{DBLP:conf/acl/GardnerC18}, which normalizes the probability across all the top-k retrieved paragraphs. This encourages the model to produce globally comparable answer scores. We denote this probability as $\mathrm{P}_{\phi}^{sn}(s|p,q)$ in contrast to the original formulation $\mathrm{P}_{\phi}(s|p,q)$ that normalizes the probability within each paragraph.

\subsection{The Pretrainining Method} 

We now describe how to pretrain the retrieval module using a better data generation strategy and a progressive training paradigm. Figure~\ref{fig:pretrain_method} depicts the whole pretraining process.

\paragraph{Pretraining Data Generation} Previous dense retrieval approaches usually rely on simple heuristics to generate synthetic matching pairs for pretraining, which do not necessarily reflect the underlying matching pattern between questions and paragraphs.
To minimize the gap between pretraining and the end task, we learn to generate high-quality questions from the paragraphs using a state-of-the-art pretrained seq2seq model, \emph{i.e.}, BART~\cite{lewis2019bart}.
More specifically, we finetune BART on the original NaturalQuestions dataset~\cite{DBLP:journals/tacl/KwiatkowskiPRCP19} such that it learns to generate questions given the groundtruth answer string and the groundtruth paragraph (labeled as \textit{long answer} in NaturalQuestions). 
We concatenate the paragraph and the answer string with a separating token as the input to the BART model. We find this simple input scheme is effective enough to generate high-quality questions, achieving a 55.6 ROGUE-L score on the dev set. Samples of the generated questions could be found in the appendix.
Afterward, we use spaCy\footnote{\url{https://spacy.io}} to recognize potential answer spans (named entities or dates) in all paragraphs in the corpus and use the finetuned BART model to generate the questions conditioned on the paragraph and each of the potential answers. 

It is worth noting that the groundtruth answer paragraph supervision at this step could be eventually dropped and we could just use weakly supervised paragraphs to train the question generator; thus, our system becomes fully weak-supervised. As the pretraining process takes a long training time and lots of resources, we are unable to repeat the whole pretraining process with the weakly-supervised question generator. However, additional question generation experiments suggest that while using weakly-supervised paragraphs, the question generator still generates high-quality questions, achieving a ROUGE-L score of 49.6.\footnote{For reference, a state-of-the-art QG model~\cite{ma2019improving} trained with strong supervision achieves 49.9 ROUGE-L on a similar QA dataset also collected from real-user queries.} 

\paragraph{In-batch Negative Sampling} 
To save computation and improve the sample efficiency of pretraining, we choose to use in-batch negative sampling~\cite{in-batch-negative-sampling} instead of gathering negative paragraphs for each question to pretrain the retrieval module. Specifically, for each pair $(q, p)$ within a batch $B$, the paragraphs paired with other questions are considered negative paragraphs for $q$. Thus, the pretraining objective for each generated question is to minimize the negative log-likelihood of selecting the correct $p$ among all paragraphs in the batch:
\begin{align}
    \mathcal{L}_{pre} = - \log  \mathrm{P}_{\theta}(p|q).
\end{align}
A graphic illustration of this strategy is shown in Figure~\ref{fig:pretrain_method}. As the batch size is usually very small compared to the number of all the paragraphs in the corpus, the pretraining task is much easier compared to the final retrieval task at inference time. In the whole corpus, there are usually lots of similar paragraphs and these paragraphs could act as strong distractors for each other in terms of both paragraph ranking and answer extraction. The desired retrieval model should be able to learn fine-grained matching instead of just learning to distinguish obviously different paragraphs. However, since existing dense retrieval methods typically use uniform batch sampling, there could be many easy negative samples in each batch, and they can only provide weak learning signals. Thus, a large batch size is usually adopted to include sufficient effective negative samples. Unfortunately, this is generally not applicable without hundreds of GPUs. 

\paragraph{The Progressive Training Paradigm} 

To provide effective negative samples, we propose a progressive training algorithm, as shown in the lower part of Figure~\ref{fig:pretrain_method}. The key idea is to leverage the retrieval model itself to find groups of similar paragraphs.
At a certain training step, we use the paragraph encoder at that moment to encode the whole corpus and cluster all $(q, p)$ pairs into many groups based on the similarity of their paragraph encodings. These groups are supposed to include similar paragraphs and potentially related questions. Then, we continue our pretraining by sampling each batch from one of the clusters. By doing this, we can provide challenging and effective negative paragraphs for each question, even with small batch size. Every time we recluster the whole corpus, the model will be encouraged to learn finer-grained matching between questions and paragraphs. Algorithm~\ref{alg:pretrain} provides a formal description of the entire process. Note that our training algorithm shares spirits with Curriculum Learning~\cite{bengio2009curriculum} and Self-Paced Learning~\cite{DBLP:conf/nips/KumarPK10,DBLP:conf/aaai/JiangMZSH15}, in which the models are trained with harder instances as the training progresses. Instead of utilizing a predefined or dynamically generated order of all the instances according to their easiness, our algorithm makes use of a dynamic grouping of all the training instances and is specifically designed for the efficient in-batch negative sampling paradigm.

\begin{algorithm}[t] 
\small
    \caption{The Clustering-based Progressive Pretraining}
    \label{alg:pretrain}
    \begin{algorithmic}[1]
        \State 
        \textbf{Input:}\\ 
        a) all $(q, p)$ pairs from the question generation model; \\
        b) the retrieval module $\mathrm{BERT}_{Q}$ and $\mathrm{BERT}_{P}$; 
        \While{not finished} 
        \State Encode the whole corpus with $\mathrm{BERT}_{P}$;
        \State Clustering all paragraphs into $C$ clusters using the dense encodings;
        \For{updates = 1:$K$}
        \State Random sample a paragraph cluster;
        \State Sample $B$ paragraphs from the cluster;
        \State Fetch the corresponding questions;
        \State Calculate gradients wrt $\mathcal{L}_{pre}$;
        \If {updates \% $U$ == 0
        }
        \State Update $\mathrm{BERT}_{Q}$ and $\mathrm{BERT}_{P}$;
        \EndIf
        \EndFor
        \EndWhile
    \end{algorithmic}
\end{algorithm}


\begin{table*}[!t]
    \centering
    \small
    \begin{tabular}{lccc}
    \toprule
       \multirow{2}{*}{\textbf{Method}} & \multicolumn{3}{c}{\textbf{Dataset}} \\ & NaturalQuestions-Open & WebQuestions & CuratedTREC \\
    \midrule
    DrQA \cite{chen-etal-2017-reading} & - & 20.7 & 25.7  \\
    R$^{3}$ \cite{DBLP:journals/corr/abs-1709-00023} & - & 17.1 & 28.4 \\
    DSQA \cite{lin-etal-2018-denoising} & - & 25.6 & 29.1 \\
    HardEM \cite{min-etal-2019-discrete} & 28.1 & - & - \\
    PathRetriever \cite{asai-etal-2019-path} &  32.6 & - & -  \\
    WKLM \cite{DBLP:journals/corr/abs-1912-09637} & - & 34.6 & - \\
    GraphRetriever \cite{min-etal-2019-graph} &  34.5 & 36.4 & - \\
    
    \midrule 
    ORQA \cite{lee-etal-2019-latent} & 33.3 & 36.4 & 30.1 \\
    
    \textbf{ProQA}(Ours) & \textbf{37.4} & \textbf{37.1} & \textbf{34.6} \\
    
    \bottomrule
    \end{tabular}
    \caption{Open-domain QA results in terms of exact answer match (EM). The first part of the table shows results from methods that use the traditional IR component. Note that these methods retrieve more paragraphs (typically dozens) than dense retrieval methods listed in the second part of the table, which only finds answers from the top-5.}
    \label{tab:QA}
\end{table*}

\subsection{QA Finetuning} 
Once pretrained, we use the paragraph encoder to encode the corpus into a large set of dense vectors. Following previous practice, we only finetune the question encoder and the reading module so that we can reuse the same dense index for different datasets. For every training question, we obtain the question representation $h_q$ from the question encoder and retrieve the top-k paragraphs $D^{k}$ on the fly using an existing maximum inner product search package. To train the reading module, we apply the shared-normalization trick and optimize the marginal probability of all matched answer spans in the top-k paragraphs:
\begin{align}
\label{loss:reader}
    \mathcal{L}_{reader} = -\log{\sum_{p\in D^{\ast}}}\sum_{s\in S^{\ast}_{p}}\mathrm{P}_{\phi}^{sn}(s|p,q).
\end{align}
In additional to the reader loss, we also incorporate the ``early" loss used by \citet{lee-etal-2019-latent}, which updates the question encoder using the top-5000 dense paragraph vectors. If we define $D^{*}_{5000}$ as those paragraphs in the top-5000 that contain the correct answer, then the ``early" loss is defined as:

\begin{align}
\label{loss:early}
    \mathcal{L}_{early} = -\log{\sum_{p\in D^{\ast}_{5000}}}\mathrm{P}_{\theta}(p|q).
\end{align}
Thus our total finetuning loss is $\mathcal{L}_{early} + \mathcal{L}_{reader}$. Note this is different from the joint formulation used by \citet{lee-etal-2019-latent} and \citet{guu2020realm}, which consider the paragraphs as latent variables when calculating $\mathrm{P}(s|q)$. We find the joint objective does not bring additional improvements, especially after we use shared normalization. More variants of the finetuning objectives will be discussed in \S\ref{sec:ft_analsis}. At inference time, we use a linear combination of the retrieval score and the answer span score to rank the answer candidates from the top-5 retrieved paragraphs. The linear combination weight is selected based on the validation performance on each tested dataset.

\section{Experiments}

\subsection{Datasets}
We center our studies on QA datasets that reflect real-world information-seeking scenarios. We consider \textbf{1) NaturalQuestions-Open}~\cite{DBLP:journals/tacl/KwiatkowskiPRCP19,lee-etal-2019-latent}, which includes around 10K real-user queries (79,168/8,757/3,610 for train/dev/test) from Google Search; \textbf{2) WebQuestions}~\cite{DBLP:conf/emnlp/BerantCFL13}, which is originally designed for knowledge base QA and includes 5,810 questions (3,417/361/2,032 for train/dev/test) generated by Google Suggest API; \textbf{3) CuratedTREC}~\cite{DBLP:conf/clef/BaudisS15}, which includes 2,180 real-user queries (1,353/133/694 for train/dev/test) from MSNSearch and AskJeeves logs. Compared to other datasets such as SQuAD~\cite{DBLP:conf/emnlp/RajpurkarZLL16} and TriviaQA~\cite{DBLP:conf/acl/JoshiCWZ17}, questions in these datasets are created without the presence of ground-truth answers and the answer paragraphs, thus are less likely to have lexical overlap with the paragraph.


\subsection{Essential Implementation Details}
\label{implementation_details}
For pretraining, we use a batch size of 80 and accumulate the gradients every 8 batches. We use the Adam optimizer~\cite{kingma2014adam} with learning rate 1e-5 and conduct 90K parameter updates. Following previous work~\cite{lee-etal-2019-latent}, we use the 12-20-2018 snapshot of English Wikipedia as our open-domain QA corpus. When splitting the documents into chunks, we try to reuse the original paragraph boundaries and create a new chunk every time the length of the current one exceeds 256 tokens. Overall, we created 
12,494,770 text chunks, which is on-par with the number (13M) reported in previous work. These chunks are also referred to as paragraphs in our work. For progressive training, we recluster all the chunks with k-means around every 20k updates using the paragraph encodings. 

While finetuning the modules for QA, we fix the paragraph encoder in the retrieval module. For each question, we use the top-5 retrieved paragraphs for training and skip the question if the top-5 paragraphs fail to cover the answer. The MIPS-based retrieval is implemented with FAISS~\cite{DBLP:journals/corr/JohnsonDJ17}. On NaturalQuestions-Open, we finetune for 4 epochs. To save the finetuning time on this large dataset, we only use a subset (2,000 out of 8,757) of the original development set for model selection. For WebQuestions and CuratedTREC (both of them are much smaller), we finetune for 10 epochs. The optimizer settings are consistent with the pretraining. Hyperparameters and further details can be found in the appendix.

\subsection{QA Performance}

Following existing studies, we use the exact match (EM) as the evaluation metric, which indicates the percentage of the evaluation samples for which the predicted span matches the groundtruth answers. In Table~\ref{tab:QA}, we first show that our progressive method (denoted as \textbf{ProQA}) is superior to all of the open-domain QA systems (the upper part of the table) that use conventional IR methods, even though we only use the top-5 paragraphs to predict the answer while these methods use dozens of retrieved paragraphs. For the dense retrieval methods, we compare with ORQA~\cite{lee-etal-2019-latent}, which is most relevant to our study but simply uses pseudo question-paragraph pairs for pretraining and also requires a larger batch size (4,096). We achieve much stronger performance than ORQA with much fewer updates and a limited number of GPUs. To the best of our knowledge, this is the first work showing that an effective dense corpus index can be obtained without using highly expensive computational resources. The reduced requirement of computation also makes our method easier to replicate for corpora in different domains. 

\begin{table}[t]
    \centering
    \small
    \begin{tabular}{ccccc}
    \toprule
        Method & EM & model size & batch size & \# updates\\
        \midrule
        ORQA & 33.3 & 330M & 4096 & 100K \\
        T5 & 36.6 & 11318M & - & - \\
        REALM &  \textbf{40.4} & 330M & 512 & 200K\\
        \midrule 
        \textbf{ProQA} & 37.4 & 330M & 80*8 & 90K \\
         
    \bottomrule
    \end{tabular}
    \caption{Resource comparison with SOTA models. EM scores are measured on NaturalQuestions-Open. \textit{batch size} and \textit{updates} all refer to the dense index pretraining. Note that REALM uses ORQA to initialize its parameters and we only report the numbers after ORQA initialization. ``80*8" indicates that we use a batch size of 80 and accumulate the gradients every 8 batches.}
    \label{tab:resource}
\end{table}

In Table~\ref{tab:resource}, we compare our method other more recently published QA systems in terms of both performance and computation cost. It is worth noting that although we need to use a BART model to generate training questions for Wikipedia documents, the inference cost of the question generator is still much lower than the training cost of our system and is not significant for comparing the overall computation cost: with the same GPU hardware, generating all the questions takes less than 1/6 of the training time. When using the $\textit{batch size} \times \# \textit{ of updates}$ to approximate the training FLOPs, we see that our system is at least 6 times more efficient than ORQA. Compared to the recent proposed T5~\cite{roberts2020} approach, which converts the QA problem into a sequence-to-sequence (decode answers after encoding questions) problem and relies on the large model capacity to answer questions without retrieving documents, our system achieves better performance and is also much faster at inference time, due to the much smaller model size. The state-of-the-art REALM model~\cite{guu2020realm} uses a more complicated pretraining approach that requires asynchronously refreshing the corpus index at train time. As it relies on ORQA initialization and further pretraining updates, it is even more computational expensive at training time. Also, as our method directly improves the ORQA pretraining, our method could easily stack with the REALM pretraining approach. 

Concurrent to our work, \citet{karpukhin2020dense} show that it is possible to use the groundtruth answer paragraphs in the original NaturalQuestions dataset to train a stronger dense retriever. However, they use a larger index dimension (768) while encoding paragraphs and also retrieve more paragraphs (20$\sim$100) for answer extraction. As a larger index dimension naturally leads to better retrieval results~\cite{luan2020sparse} (despite sacrificing search efficiency) and using more paragraphs increases the recall of matched answer spans\footnote{While using more paragraphs, we achieve 40.6 EM (compared to 41.5 EM in \cite{karpukhin2020dense}) even with a much smaller index dimension. Also, we did not use the gold paragraphs (strong supervision) to train our reading module.}, this concurrent result is not directly comparable to ours, and we leave the combination effect of efficient pretraining and strong supervision (\emph{i.e.}, using human-labeled paragraphs) to future work.
\subsection{Ablation Studies}

\begin{table}[t]
    \centering
    \small
    \begin{tabular}{lcccc}
    \toprule
        Method & R@5 & R@10 & R@20\\
    \midrule
        ProQA (90k) & \textbf{52.0} & \textbf{61.0} & \textbf{68.8} & \\
        ORQA$^\star$ (90k) & 20.4 & 29.0 & 37.2 \\
        ProQA (no clustering, 90k) & 42.9 & 52.6 & 60.8\\
        ProQA (no clustering; 70k) & 43.8 & 53.5 & 61.3 \\
        ProQA (no clustering; 50k) & 38.8 & 48.2 & 56.7 \\
    \bottomrule
    \end{tabular}
    \caption{Ablation studies on different pretraining strategies. The retrieval modules (Recall@k) are tested on WebQuestions. $^\star$Our reimplementation.}
    \label{tab:retrieve_ablation}
\end{table}

To validate the sample efficiency of our method, we replicate the inverse-cloze pretraining approach from ORQA using the same amount of resource as we used while training our model, \emph{i.e.}, the same batch size and updates (90k). We also study the effect of the progressive training paradigm by pretraining the model with the same generated data but without the clustering-based sampling. We test the retrieval performance on the WebQuestions test set before any finetuning. We use Recall@k as the evaluation metric, which measures how often the answer paragraphs appear in the top-k retrieval. The results are shown in Table~\ref{tab:retrieve_ablation}. 
We can see that for the non-clustering version of our method, the improvements are diminishing as we reach certain training steps, while the progressive training algorithm brings around $8\%$ improvements on different retrieval metrics. This suggests the importance of introducing more challenging negative examples in the batch when the batch size is limited. Comparing the no-clustering version of our method against ORQA, we see that using our data generation strategy results in much better retrieval performance (more than $22\%$ improvements on all metrics). 

\subsection{Analysis on Finetuning Objectives}
\label{sec:ft_analsis}
Noting that different finetuning configurations have been used in existing studies , we conduct additional experiments to investigate the efficacy of different finetuning objectives and provide insights for future research that intends to focus on improving the model itself. Specifically, we study the effects of using a joint objective~\cite{lee-etal-2019-latent,guu2020realm} and adding an additional reranking objective that is commonly used in sparse-retrieval QA systems~\cite{min-etal-2019-discrete,DBLP:journals/corr/abs-1912-09637}. 

The joint objective treats the retrieved paragraphs as latent variables and optimizes the marginal probability of the matched answer spans in all paragraphs:
\begin{align}
\label{loss:joint}
    \mathcal{L}_{joint} = -\log{\sum_{p\in D^{\star}}\mathrm{P}_{\theta}(p|q)}\sum_{s\in S^{\ast}_{p}}\mathrm{P}_{\phi}(s|p,q).
\end{align}
The reranking objective is usually implemented through a paragraph reranking module that uses a question-dependent paragraph encoder, \emph{e.g.}, a BERT encoder that takes the concatenation of the question and paragraph as input.
This kind of reranker has been shown to be beneficial to conventional IR methods since it can usually provide more accurate paragraph scores than the TF-IDF or BM25 based retriever while ranking the answer candidates. 
To implement this reranking module, we simply add another reranking scoring layer to our BERT-based span prediction module $\mathrm{P}_{\phi}(
s|p,q)$, which encodes the paragraphs in a question-dependent fashion. At inference time, we use the paragraph scores predicted by this reranking component instead of the pretrained retrieval model to guide our final answer selection. 

\begin{table}[!t]
    \centering
    \small
    \begin{tabular}{ccccc}
    \toprule
        \multirow{2}{*}{id} & \multicolumn{3}{c}{\textbf{Objective Settings}} & \multirow{2}{*}{EM}   \\
        & \textit{joint} & \textit{rerank} & \textit{shared-norm} & \\
        \midrule
        1 &- & - & $\checkmark$ & 38.5\\
        2 & $\checkmark$ & - &  $\checkmark$ & 38.3\\
        3 & - & $\checkmark$ &  $\checkmark$ & 38.2\\
        4 & $\checkmark$ & - & - & 36.2\\
        5  &- & - & - & 35.1\\
    \bottomrule
    \end{tabular}
    \caption{Analysis on different finetuning objectives on  NaturalQuetions-Open. EM scores are measured on the 2,000 validation samples we used for model selection.}
    \label{tab:qa_analysis}
\end{table}

Table~\ref{tab:qa_analysis} shows the results of different objective settings. Comparing the results of (4) and (5), we can see that the joint objective can bring some improvements when shared-normalization is not applied. However, it does not yield improvements when shared-normalization is applied, according to the results of (1) and (2).
By comparing (1) and (3), we see that with the strong pretrained retrieval model, adding an extra reranking module that uses question-dependent paragraph encodings is no longer beneficial. This is partially because our pretrained retrieval model gets further improved during finetuning, in contrast to a fixed TF-IDF based retriever. Finally, from (1) and (5), we see that the shared normalization brings much larger improvements than the other factors. This aligns with the findings from an existing work~\cite{DBLP:conf/emnlp/WangNMNX19} that only tested on SQuAD questions.

\section{Error Analysis}
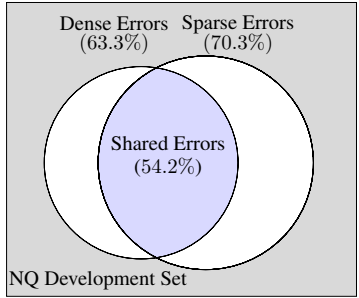
\begin{figure}
\centering
\resizebox{0.3\textwidth}{!}{
\begin{tikzpicture}
\def\bigrectangle{(-2.5,-2.5) rectangle (4,3)}
\def\firstcircle{(0,0) circle (1.8)}
\def\secondcircle{(1.2,0) circle (2)}

\scope
    \clip \firstcircle \secondcircle \bigrectangle;
    \fill[gray!30!white] \bigrectangle;
\endscope

\scope
    \clip \secondcircle;
    \fill[blue!15!white] \firstcircle;
\endscope

\draw \firstcircle (-0.5,2.35) node [text=black,above] {Dense Errors}
    \firstcircle (-0.5,1.9) node [text=black,above] {$(63.3\%)$}
      \secondcircle (1.8,2.28) node [text=black,above] {Sparse Errors}
      \secondcircle (1.8,1.9) node [text=black,above] {$(70.3\%)$}
      \bigrectangle (-0.8,-2.5) node [text=black,above] {NQ Development Set}
      \secondcircle (0.5,0.1) node [text=black,above] {Shared Errors}
      \secondcircle (0.5,-0.4) node [text=black,above] {($54.2\%$)};

\end{tikzpicture}
}
    \caption{The error sets (from the official dev set of NaturalQuesitons-Open) of our dense retrieval method and a baseline system using sparse retrieval methods like TF-IDF. We use circles to represent the error sets of both systems. The percentages show the relative size of each set  in terms of  the whole dev set.}
    \label{fig:error_set}
\end{figure}

To investigate the fundamental differences of the dense and sparse retrieval methods in open-domain QA, we conduct an error analysis using both the proposed method and a baseline system that uses TF-IDF and BM25 for retrieval. This baseline uses a similar retrieval pipeline as \citet{min-etal-2019-discrete} and is trained with the finetuning objective defined in Eq.~\ref{loss:reader}. This sparse retrieval baseline achieves 29.7 EM on the official dev set of NaturalQuestions-Open while our method achieves 36.7 EM\footnote{Note that this number is different from the number in Table 4 as we did not use the whole dev set for model selection.}. Figure~\ref{fig:error_set} shows the Venn diagram of the error sets from both systems. Our key findings are summarized in the following paragraphs. 

\paragraph{The difference retrieval paradigms could complement each other.} First, according to the error set differences (shown by the white regions in Figure~\ref{fig:error_set}), a considerable portion of the error cases ($9.1\%$ of the devt set) of our dense-retrieval system does not occur in the sparse-retrieval system and vice versa. This suggests the necessity of incorporating different retrieval paradigm when building real-world applications. In fact, the hybrid approach has already been adopted by a phrase-level retrieval method~\cite{DBLP:conf/acl/SeoLKPFH19} and concurrent studies~\cite{luan2020sparse,karpukhin2020dense}.


\paragraph{Both systems are underestimated.} As shown in Figure~\ref{fig:error_set}, $54.2\%$ of the questions cannot be correctly answered by either system. However, our manual inspection on 50 of the shared error cases suggests that around $30\%$ of these errors are due to annotation issues ($14\%$) or the ambiguous nature of real-user queries ($16\%$). One obvious annotation issue is the incompleteness of the answer labels. Example questions include \textit{``When did Brazil lose to in 2014 World Cup?''}, to which both \textit{``Germany"} and \textit{``Netherlands"} are correct answers. This issue occurs because the annotators of NaturalQuestions only have a local view of the knowledge source as they are only asked to label the answer span using one document. In terms of the ambiguous questions, many of them are due to constraints unspecified by the question words, such as \textit{``What is the population of New York City?"} (the time constraint is implicit) or \textit{``When did Justice League come out in Canada?''} (needs entity disambiguation). This kind of questions result from the information-seeking nature of the open-domain QA task where the users usually use the minimal number of words for searching and they are not aware of the potential ambiguous factors. To solve this kind of questions, an interactive QA system might be necessary. In the appendix, we show more ambiguous questions in which other kinds of constraints are missing. 

\section{Related Work}
The task of answering questions without specifying specific domains has been intensively studied since the earlier TREC QA competitions~\cite{DBLP:conf/trec/Voorhees99}. Studies in the early stage~\cite{DBLP:conf/www/KwokEW01,DBLP:conf/emnlp/BrillDB02,ferrucci2010building,baudivs2015yodaqa} mostly rely on highly sophisticated pipelines and heterogeneous resources. Built on the recent advances in machine reading comprehension, \citet{chen-etal-2017-reading} show that open-domain QA can be simply formulated as a reading comprehension problem with the help of a standard IR component that provides candidate paragraphs for answer extraction. This two-stage formulation is simple yet effective to achieve competitive performance while using Wikipedia as the only knowledge resource.  

Following this formulation, a couple of recent studies have proposed to improve the system using stronger reading comprehension models~\cite{DBLP:journals/corr/abs-1902-01718,DBLP:conf/iclr/WangY0ZGCWKTC18}, more effective learning objectives~\cite{DBLP:conf/acl/GardnerC18,min-etal-2019-discrete,DBLP:conf/emnlp/WangNMNX19} or paragraph reranking models~\cite{DBLP:journals/corr/abs-1709-00023,lin-etal-2018-denoising,DBLP:conf/emnlp/LeeYKKK18}. However, the retrieval components in these systems are still based on traditional inverted index methods, which are efficient but might fail when the target paragraph does not have enough lexicon overlap with the question. 

In contrast to the sparse term-based features used in TF-IDF or BM25, dense paragraph vectors learned by deep neural networks~\cite{DBLP:conf/nips/ZhangSWGHC17,DBLP:journals/corr/ConneauKSBB17} can capture much richer semantics beyond the n-gram term features. To build effective paragraph encoders tailed for the paragraph retrieval in open-domain QA, more recent studies~\cite{lee-etal-2019-latent,chang2020pre,guu2020realm} propose to pretrain Transformer encoders~\cite{DBLP:conf/nips/VaswaniSPUJGKP17} 
with objectives that simulate the semantic matching between questions and paragraphs. For instance, \citet{lee-etal-2019-latent} uses the inverse cloze pretraining task to train a bi-encoder model to match a sentence and the paragraph in which the sentence belongs to. These approaches demonstrate promising performance but require a lot of resources for pretraining. The focus of this paper is to reduce the computational requirements of building an effective corpus index. 

\section{Conclusion}
We propose an efficient method for pretraining the dense corpus index which can replace the traditional IR methods in open-domain QA systems. The proposed approach is powered by a better data generation strategy and a simple yet effective data sampling protocol for pretraining. With careful finetuning, we achieve stronger QA performance than ORQA that uses much more computational resources. We hope our method could encourage more energy-efficient pretraining methods in this direction such that the dense retrieval paradigm could be more widely used in different domains. 

\section{Acknowledgement}
The research was partly supported by the J.P. Morgan AI Research Award. The views and conclusions contained in this document are those of the authors and should not be interpreted as representing the official policies, either expressed or implied, of the funding agency.

\bibliography{eacl2021}
\bibliographystyle{acl_natbib}
\clearpage
\appendix

\section{Appendix}

\subsection{Further Implementation Details}
The pretraining process takes 4 TITAN RTX GPUs, each with a 24G memory. We use the NVIDIA Apex package for mixed-precision training. 90K parameter updates take around 7 days to finish. For QA experiments, We use the \textbf{IndexIVFFlat} index for efficient search. We assign all the vectors to 100 Voronoi cells and only search from the closest 20 cells. The random seed is set as 3 for all QA datasets. We use a batch size of 8 (8 questions, each of them are paired with 5 paragraphs) for NaturalQuestions-Open and 1 for the other datasets. We limit the maximum answer length to 10 subword tokens. For NaturalQuestions-Open, we evaluate the model every 1000 updates and save the best checkpoint based on validation EM. For WebQuesitons and CuratedTREC, we evaluate the model after every epoch. As neither of these two small datasets has an official dev set, we use a small split to find the best hyperparameters and then retrain the model with all the training questions. To accelerate training, especially for the early loss function which requires annotate the top5000 retrieved paragraphs, we pre-annotate the top10000 paragraphs retrieved by the untuned retrieval module and build an answer paragraph set for each question. At finetuning time, we direct check whether a particular paragraph is in the precomputed paragraph set, instead of doing string matching for each of the 5000 paragraphs. Our BERT implementations are based on huggingface Transformers\footnote{\url{https://github.com/huggingface/transformers}}.

\subsection{Qualitative Examples}
Here we include more examples that complement the results and analysis of the paper. Table~\ref{Question_Examples} shows the generated questions from the finetuned BART model and Table~\ref{error_cases} complements the error analysis.

\begin{table*}[h]
\centering
\small
\begin{tabular}{l}
\toprule
\multicolumn{1}{p{14cm}}{\textbf{Gold Paragraph:}  ``Does He Love You'' is a song written by Sandy Knox and Billy Stritch, and recorded as 
a duet by American country music artists Reba McEntire and \textcolor{red}{\underline{Linda Davis}}. It was released in August 1993 as the first single from Reba's album Greatest Hits Volume Two. It is one of country music 's several songs about a love triangle.} \\
\textbf{Original Question:} Who sings does he love me with reba? \\ 
\textbf{Generated Question:} Who sings with reba mcentire on does he love you? \\
\midrule
\multicolumn{1}{p{14cm}}{\textbf{Gold Paragraph:} Invisible Man First edition Author Ralph Ellison Country United States Language English Genre Bildungsroman African-American literature social commentary Publisher Random House Publication date 1952 Media type Print (hardcover and paperback) Pages \textcolor{red}{\underline{581}} (second edition) ...} \\
\textbf{Original Question:} How many pages is invisible man by ralph ellison? \\ 
\textbf{Generated Question:} How many pages in the invisible man by ralph ellison? \\
\midrule
\multicolumn{1}{p{14cm}}{\textbf{Gold Paragraph: }The Great Lakes (French: les Grands-Lacs), also called the Laurentian Great Lakes and the Great Lakes of North America, are a series of interconnected freshwater lakes located primarily in the upper mid-east region of North America, on the Canada--United States border, which connect to the Atlantic Ocean through \textcolor{red}{\underline{the Saint Lawrence River}}. They consist of Lakes Superior, Michigan, Huron (or Michigan--Huron), Erie, and Ontario.}\\
\textbf{Original Question:} Where do the great lakes meet the ocean? \\ 
\textbf{Generated Question:} Where do the great lakes of north america meet the atlantic? \\
\midrule
\multicolumn{1}{p{14cm}}{\textbf{Gold Paragraph: } My Hero Academia: Two Heroes~, Hepburn:Boku no Hiro Academia THE MOVIE: Futari no Hiro) is a 2018 Japanese anime superhero film based on the manga My Hero Academia by Kohei Horikoshi. Set between the second and third seasons of the anime series, the film was directed by Kenji Nagasaki and produced by Bones. Anime Expo hosted the film's world premiere on \textcolor{red}{\underline{July 5, 2018}}, and it was later released to theaters in Japan on August 3, 2018.}\\
\textbf{Original Question:} When does the new my hero academia movie come out? \\ 
\textbf{Generated Question:} When does the my hero academia two heroes movie come out? \\
\midrule
\multicolumn{1}{p{14cm}}{\textbf{Gold Paragraph: } Victoria's Secret Store, 722 Lexington Ave, New York, NY Type Subsidiary Industry Apparel Founded June 12, 1977; 40 years ago (1977-06-12 ) Stanford Shopping Center, Palo
 Alto, California, U.S. Founder \textcolor{red}{\underline{Roy Raymond}} Headquarters Three Limited Parkway, Columbus , Ohio , U.S. Number of locations 1,017 company - owned stores 18 independently owned stores Area served ...} \\
 \textbf{Original Question:} Who was the creator of victoria's secret?\\
 \textbf{Generated Question:} Who is the founder of victoria's secret and when was it founded?\\
\bottomrule
\end{tabular}
\caption{Samples of the generated questions. The answer spans are underlined. Here we show the generated questions for samples at the beginning of the official NaturalQuestions-Open dev data. We only skip the samples whose gold paragraphs are not natural paragraphs (\emph{e.g.}, incomplete sentences). }
\label{Question_Examples}
\end{table*}

\begin{table*}[h]
\centering
\small
\begin{tabular}{l}
\toprule
\multicolumn{1}{p{7cm}}{\textbf{Question: }  What is a ford mondeo in the usa?} \\ \textbf{Annotated Answers: } ford contour, mercury mystique, ford fusion\\
\textcolor{red}{\textit{ambiguous;}} could be asking about a particular car type (mid-sized car) instead of brand series\\
\midrule
\multicolumn{1}{p{7cm}}{\textbf{Question: }  air flow in the eye of a hurricane?} \\ \textbf{Annotated Answers: } no wind\\
\textcolor{red}{\textit{ambiguous;}} question itself is hard to understand\\
\midrule
\multicolumn{1}{p{14cm}}{\textbf{Question: }  Who wrote I'll be there for you?} \\ \textbf{Annotated Answers: } Michael Skloff, Marta Kauffman, Allee Willis, David Crane, Phil Solem, Danny Wilde, The Rembrandts\\
\textcolor{red}{\textit{ambiguous;}} there are multiple songs having this name\\
\midrule
\multicolumn{1}{p{7cm}}{\textbf{Question: }  Where do you go for phase 1 training?} \\ \textbf{Annotated Answers: } army foundation college\\
\textcolor{red}{\textit{ambiguous;}} the meaning of phase 1 is vague, could have different meanings in different context\\
\midrule
\multicolumn{1}{p{14cm}}{\textbf{Question: }  When does the new spiderman series come out?} \\ \textbf{Annotated Answers: } August 19 , 2017\\
\textcolor{red}{\textit{ambiguous;}} time constraint missing\\
\midrule
\multicolumn{1}{p{14cm}}{\textbf{Question: } Where did the super bowl take place this year?} \\ \textbf{Annotated Answers: } minneapolis, minnesota\\
\textcolor{red}{\textit{ambiguous;}} the year cannot be inferred from the question words alone\\
\midrule
\end{tabular}
\caption{Error cases that include ambiguous questions.}
\label{error_cases}
\end{table*}

\end{document}